\documentclass[runningheads]{llncs}
\pdfoutput=1
\usepackage{graphicx}
\usepackage{tikz}
\usepackage{comment}
\usepackage{amsmath,amssymb} % define this before the line numbering.
\usepackage{color}
\usepackage{booktabs}
\usepackage{tabularx}
\usepackage{multirow}
\usepackage{caption}
\usepackage{subcaption}
\usepackage{makecell}
\usepackage{bbding}
\usepackage[accsupp]{axessibility}  % Improves PDF readability for those with disabilities.

\begin{document}
% \linenumbers
\pagestyle{headings}
\mainmatter
\def\ECCVSubNumber{6420}  % Insert your submission number here

\title{A Codec Information Assisted Framework for Efficient Compressed Video Super-Resolution} % Replace with your title

\begin{comment}
\titlerunning{ECCV-22 submission ID \ECCVSubNumber} 
\authorrunning{ECCV-22 submission ID \ECCVSubNumber} 
\author{Anonymous ECCV submission}
\institute{Paper ID \ECCVSubNumber}
\end{comment}
%******************

%\begin{comment}
\titlerunning{Codec Information Assisted Video Super-Resolution}
% an abbreviated paper title here\orcidlink{0000-0001-6738-3462}\orcidlink{0000-0002-7124-5182}
%
\author{Hengsheng Zhang\inst{1}\index{Zhang, Hengsheng} \and
Xueyi Zou\inst{2}\index{Zou, Xueyi} \and
Jiaming Guo\inst{2}\index{Guo, Jiaming} \and
Youliang Yan\inst{2}\index{Yan, Youliang} \and
Rong Xie\inst{1}\index{Xie, Rong} \and
Li Song\inst{1,3}\textsuperscript{\Envelope}\index{Song, Li}}
\authorrunning{Hengsheng Zhang et al.}
\institute{Institute of Image Communication and Network Engineering, Shanghai Jiao Tong University \and
Huawei Noah's Ark Lab \and MoE Key Lab of Artifical Intelligence, AI Institute, Shanghai Jiao Tong University
\email{\{hs\_zhang,xierong,song\_li\}@sjtu.edu.cn}
\email{\{zouxueyi,guojiaming5,yanyouliang\}@huawei.com}
}
%\end{comment}
%******************

\maketitle

\begin{abstract}
Online processing of compressed videos to increase their resolutions attracts increasing and broad attention. Video Super-Resolution (VSR) using recurrent neural network architecture is a promising solution due to its efficient modeling of long-range temporal dependencies. However, state-of-the-art recurrent VSR models still require significant computation to obtain a good performance, mainly because of the complicated motion estimation for frame/feature alignment and the redundant processing of consecutive video frames. In this paper, considering the characteristics of compressed videos, we propose a Codec Information Assisted Framework (CIAF) to boost and accelerate recurrent VSR models for compressed videos. Firstly, the framework reuses the coded video information of Motion Vectors to model the temporal relationships between adjacent frames. Experiments demonstrate that the models with Motion Vector based alignment can significantly boost the performance with negligible additional computation, even comparable to those using more complex optical flow based alignment. Secondly, by further making use of the coded video information of Residuals, the framework can be informed to skip the computation on redundant pixels. Experiments demonstrate that the proposed framework can save up to 70\% of the computation without performance drop on the REDS4 test videos encoded by H.264 when CRF is 23.
\end{abstract}

\section{Introduction}
Compressed videos are prevalent on the Internet, ranging from movies, webcasts to user-generated videos, most of which are of relatively low resolutions and qualities. Many terminal devices, such as smartphones, tablets, and TVs, come with a 2K/4K or even 8K definition screen. Thus, there is an urgent demand for such devices to be able to online super-resolve the low-resolution videos to the resolution of the screen definition. Video Super-Resolution (VSR) increases the video frames’ resolution by exploiting redundant and complementary information along the video temporal dimension. With the wide use of neural networks in computer vision tasks, on the one hand, neural network based VSR methods outperform traditional ones. But on the other hand, they require a lot of computation and memory, which current commercial terminal devices cannot easily provide. 

Most neural network based VSR models come with a lot of repeated computation or memory consumption. For example, sliding-window based VSR models\cite{DBLP:conf/cvpr/HarisSU19,DBLP:conf/cvpr/TianZ0X20,DBLP:conf/cvpr/WangCYDL19,DBLP:journals/tip/ChenYWSHW21} have to extract the features of adjacent frames repeatedly. Although this process can be optimized by preserving the feature maps of previous frames, it increases memory consumption. Besides, to make the most of adjacent frames’ information, frame alignment is an essential part of many such models, which is usually implemented by optical flow prediction\cite{DBLP:conf/cvpr/RanjanB17,DBLP:conf/cvpr/SunY0K18}, deformable convolution\cite{DBLP:conf/iccv/DaiQXLZHW17,DBLP:conf/cvpr/ZhuHLD19}, attention/correlation\cite{DBLP:conf/eccv/LiTGQLJ20}, and other complicated modules\cite{DBLP:conf/cvpr/JoOKK18,DBLP:conf/iccv/YiWJJ019}. This frame alignment process also increases model complexity, and many of the operators are not well supported by current terminal chipsets. 

Many VSR methods use recurrent neural networks to avoid repeated feature extraction and to exploit long-range dependencies. The previous frame's high-resolution information (image or features) is reused for the current frame prediction. Several information propagation schemes have been proposed, such as unidirectional propagation\cite{DBLP:conf/cvpr/SajjadiVB18,DBLP:conf/iccvw/FuoliGT19,DBLP:conf/eccv/IsobeJGLWT20}, bidirectional propagation\cite{DBLP:conf/cvpr/ChanWYDL21,DBLP:journals/corr/abs-2105-01237}, and the more complex grid propagation\cite{DBLP:journals/corr/abs-2104-13371,DBLP:journals/corr/abs-2103-15683}. As expected, the more complex the propagation scheme is, the better the super-resolution performs in terms of PSNR/SSIM or visual quality. However, considering the stringent computational budget of terminal devices and the online processing requirement, most complex propagation schemes, such as bidirectional propagation and grid propagation, are not good choices. Unidirectional recurrent models seem to be good candidates, but to get better performance, frame/feature alignment is also indispensable. As mentioned above, mainstream methods for alignment are computationally heavy and not well supported by current terminal chipsets.

Compared with raw videos, compressed videos have some different characteristics. When encoding, the motion relationships of the current frame and a reference frame (e.g. the previous frame) are calculated as \textbf{Motion Vectors} (MVs). The reference frame is then warped according to MVs to get the predicted image of the current time step. The differences between the predicted image and current frame are calculated as \textbf{Residuals}. MVs and Residuals are encoded in the video streams, with MVs providing motion cues of video frames and Residuals indicating the motion-compensated differences between frames. When decoding, MVs and Residuals are extracted to rebuild the video frames sequentially based on the previous rebuilt frames.

By leveraging the characteristics of compressed videos, we propose a Codec Information Assisted Framework (CIAF) to improve the performance and the efficiency of unidirectional recurrent VSR methods. To align the features of previous frame, we reuse the MVs to model the temporal relationships between adjacent frames. The models using MV-based alignment can significantly boost the performance with negligible additional computation, even reaching a comparable performance with those using more complex optical flow based alignment. To further reduce terminal device computation burden, we apply most computation (convolutions) only to changed regions of consecutive frames. For the rest areas, we reuse features of the previous frame by warping part of the feature maps generated in the last step according to MVs. The way to determine where the change happens is based on Residuals, i.e., only pixels with Residuals not equal to zero are considered to be changed. Due to the high degree of similarity between video frames, the proposed approach can skip lots of computation. The experiments show up to 70\% of computation can be saved without performance drop on the REDS4 \cite{DBLP:conf/cvpr/WangCYDL19} test videos encoded by H.264 when CRF is 23.

The contributions of this paper can be summarized as follows. (1) We propose to reuse the coded video information of MVs to model temporal relationships between adjacent frames for frame/feature alignment. Models with MV-based alignment can significantly boost performance with minimal additional computation, even matching the performance of optical flow based models. (2) We find that the coded information of Residuals can inform the VSR models to skip the computation on redundant pixels. The models using Residual-informed sparse processing can save lots of computation without a performance drop. (3) We disclose some of the crucial tricks to train the CIAF, and we evaluate some of the essential design considerations contributing to the efficient compressed VSR model.
\section{Related Work}
\label{sec:related work}
In this section, we first review the CNN-based video super-resolution work. Then, we discuss adaptive CNN acceleration techniques related to our work.
\subsection{Video Super-Resolution}
Video super-resolution (VSR) is challenging because complementary information must be aggregated across misaligned video frames for restoration. There are mainly two forms of VSR algorithms: sliding-window methods and recurrent methods.

\noindent{\bf Sliding-window methods.} Sliding-window methods restore the target high-resolution frame from the current and its neighboring frames. \cite{DBLP:conf/cvpr/CaballeroLAATWS17,DBLP:journals/ijcv/XueCWWF19} align the neighboring frames to the target frame with predicted optical flows between input frames. Instead of explicitly aligning frames, RBPN\cite{DBLP:conf/cvpr/HarisSU19} treats each context frame as a separate source of information and employs back-projection for iterative refining of target HR features. DUF\cite{DBLP:conf/cvpr/JoOKK18} utilizes generated dynamic upsampling filters to handle motions implicitly. Besides, deformable convolutions (DCNs)\cite{DBLP:conf/iccv/DaiQXLZHW17,DBLP:conf/cvpr/ZhuHLD19} are introduced to express temporal relationships. TDAN\cite{DBLP:conf/cvpr/TianZ0X20} aligns neighboring frames with DCNs in the feature space. EDVR\cite{DBLP:conf/cvpr/WangCYDL19} uses DCNs on a multi-scale basis for more precise alignment. MuCAN\cite{DBLP:conf/eccv/LiTGQLJ20} searches similar patches around the target position from neighboring frames instead of direct motion estimation. \cite{DBLP:journals/tip/ChenYWSHW21} extracts Motion Vectors from compressed video streams as motion priors for alignment and incorporates coding priors into modified SFT blocks\cite{DBLP:conf/cvpr/WangYDL18} to refine the features from the input LR frames. These methods can produce pleasing results, but they are challenging to be applied in practice on the terminal devices due to repeated feature extraction or complicated motion estimation.  

\noindent{\bf Recurrent methods.} Unlike sliding-window methods, recurrent methods take the output of the past frame processing as a prior input for the current iteration. So the recurrent networks are not only efficient but also can take account of long-range dependencies. In unidirectional recurrent methods FRVSR\cite{DBLP:conf/cvpr/SajjadiVB18}, RLSP\cite{DBLP:conf/iccvw/FuoliGT19} and RSDN\cite{DBLP:conf/eccv/IsobeJGLWT20}, information is sequentially propagated from the first frame to the last frame, so this kind of scheme has the potential to be applied for online processing. Besides, FRVSR\cite{DBLP:conf/cvpr/SajjadiVB18} aligns the past predicted HR frame with optical flows for the current iteration. RLSP\cite{DBLP:conf/iccvw/FuoliGT19} and RSDN\cite{DBLP:conf/eccv/IsobeJGLWT20} employs high-dimensional latent states to implicitly transfer temporal information between frames. Different from unidirectional recurrent networks, BasicVSR\cite{DBLP:conf/cvpr/ChanWYDL21} proposes a bidirectional propagation scheme to better exploit temporal features. BasicVSR++\cite{DBLP:journals/corr/abs-2104-13371} redesigns BasicVSR by proposing second-order grid propagation and flow-guided deformable alignment. Similar with BasicVSR++, \cite{DBLP:journals/corr/abs-2103-15683} employs complex grid propagation to boost the performance. COMISR\cite{DBLP:journals/corr/abs-2105-01237} applies a bidirectional recurrent model to compressed video super-resolution and uses a CNN to predict optical flows for alignment. Although they can achieve state-of-the-art performance, the complicated information propagation scheme and complex motion estimation make them unpractical to apply to the terminal device with online processing.
\subsection{Adaptive Inference}
Most of the existing CNN methods treat all regions in the image equally. But the flat area is naturally easier to process than regions with textures.  Adaptive inference can adapt the network structure according to the characteristics of the input. BlockDrop\cite{DBLP:conf/cvpr/WuNKRDGF18} proposes to dynamically pick which deep network layers to run during inference to decrease overall computation without compromising prediction accuracy. ClassSR\cite{DBLP:conf/cvpr/KongZ0D21} uses a ``class module" to decompose the image into sub-images with different reconstruction difficulties and then applies networks with various complexity to process them separately. Liu et al. \cite{DBLP:conf/eccv/LiuZHZZ20} establishes adaptive inference for SR by adjusting the number of convolutional layers used at various locations. Wang et al. \cite{DBLP:journals/corr/abs-2006-09603} locate redundant computation by predicted spatial and channel masks and use sparse convolution to skip redundant computation. The image-based acceleration algorithms follow the internal characteristics of images, so they can only reduce spatial redundancy.

Most of the time, the changes between consecutive frames in a video are insignificant. Based on this observation, Skip-Convolutions\cite{DBLP:conf/cvpr/HabibianACB21} limits the computation only to the regions with significant changes between frames while skipping the others. But this model is primarily applicable to high-level tasks. FAST\cite{DBLP:conf/cvpr/ZhangS17}, the most similar work with ours, employs SRCNN\cite{DBLP:conf/eccv/DongLHT14} to only generate the HR image of the first frame in a group of frames. In the following iterations, the HR blocks of the last frame are transferred to the current frame according to MVs. Finally, the up-sampled Residuals are added to the transferred HR image to generate the HR output of the current frame. The operations are on the pixel level, which can easily lead to errors. Instead of directly reusing the HR pixels from past frames, we utilize MVs to conduct an efficient alignment for unidirectional recurrent VSR systems. And the Residuals are used to determine the locations of redundancy.
\begin{figure*}[t]
	\centering
	\includegraphics[width=0.9\linewidth]{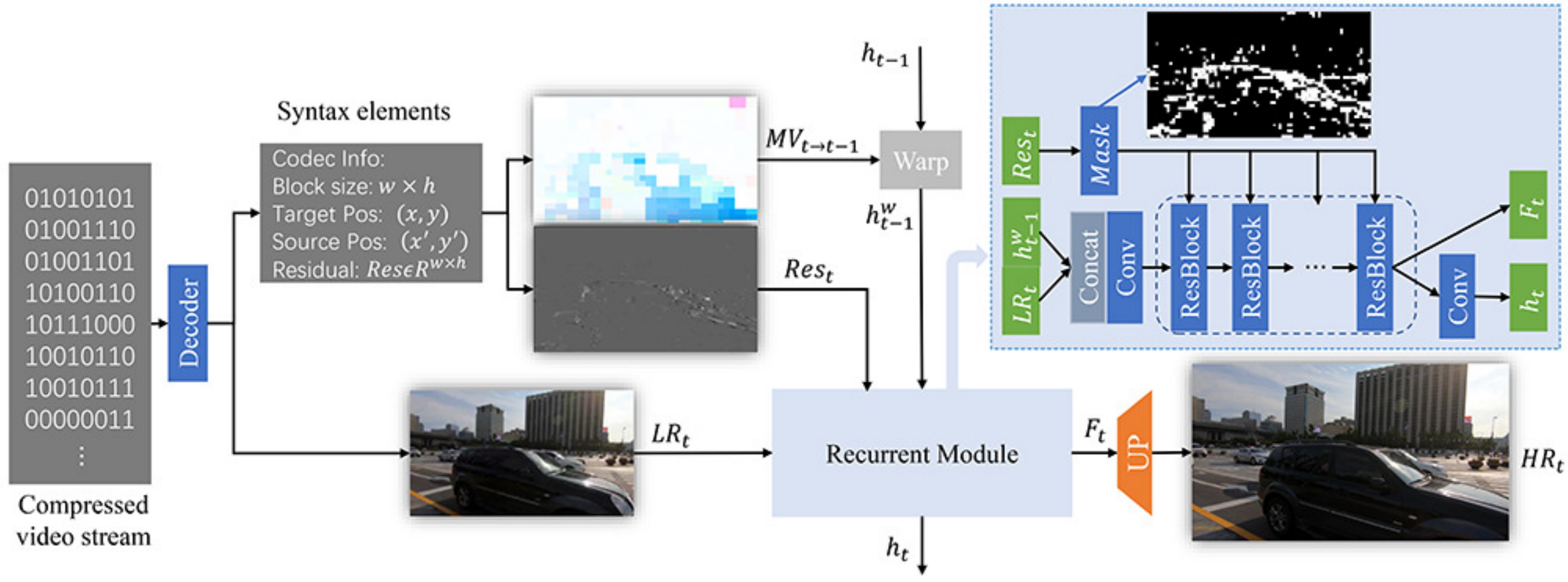}
	\caption{Overview of the proposed codec information assisted framework (CIAF). The $h_{t-1}$ is the refined features from past frame $LR_{t-1}$. Motion Vector ($MV_{t\rightarrow t-1}$) and Residuals ($Res_t$) are the codec information. In our model, we utilize the Motion Vector to align the features from the past frame. Besides, the sparse processing is applied in the Resblocks only to calculate the regions with Residuals.}
	\label{fig:MV model}
\end{figure*}
\section{Codec Information Assisted Framework}
\label{sec: method}
In this section, we first introduce the basics of video coding related to our framework. Then we present our codec information assisted framework (CIAF, Fig. \ref{fig:MV model}) consisting of two major parts, i.e., the Motion Vector (MV) based alignment and Residual informed sparse processing.
\subsection{Video coding Basics}
\begin{figure*}[h]
	\centering 
	\includegraphics[width=0.75\linewidth]{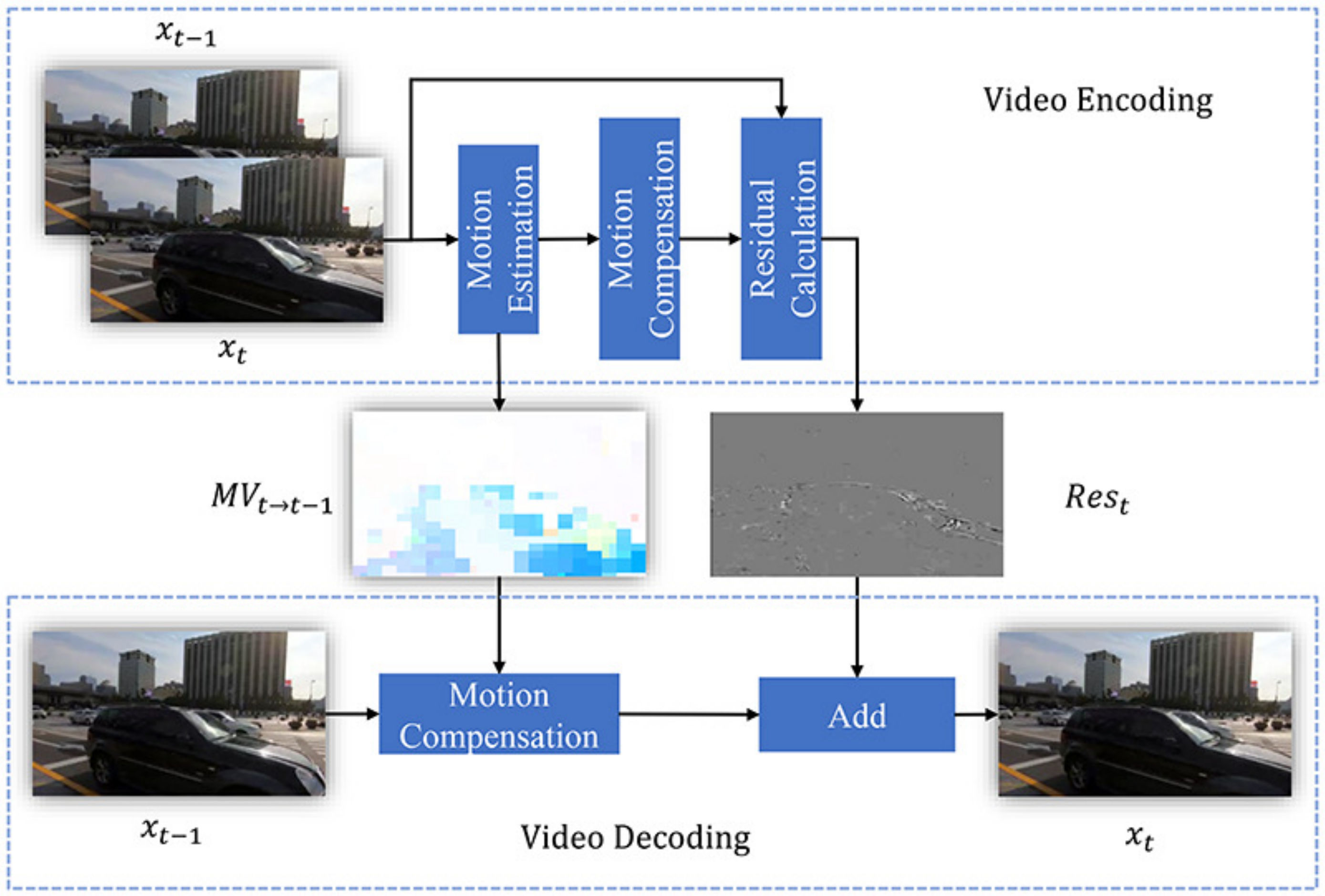}
	\caption{The Inter-Prediction Mode of video codec.}
	\label{fig:codec basics}
\end{figure*}

The Inter-Prediction Mode (Fig. \ref{fig:codec basics}) of video codec inspires our framework. Generally, there is a motion relationship between the objects in each frame and its adjacent frames. The motion relationship of this kind of object constitutes the temporal redundancy between frames. In H.264\cite{2005H}, temporal redundancy is reduced by motion estimation and motion compensation. As Fig. \ref{fig:codec basics} shows, in motion estimation, for every current block, we can find a similar pixel block as a reference in the reference frame. The relative position between the current pixel block in the current frame and the reference block in the reference frame is represented by $(MV_x, MV_y)$, a vector of two coordinate values used to indicate this relative position, known as the \textbf{Motion Vector (MV)}. In motion compensation, we use the found reference block as a prediction of the current block. Because there are slight differences between the current and reference blocks, the encoder needs to calculate the differences as \textbf{Residual}. When decoding, we first use the decoded reference frame and MVs to generate the prediction image of the target frame. Then we add decoded Residuals to the prediction image to get the target frame. In our paper, we reuse the MVs and Residuals to increase the efficiency of unidirectional recurrent VSR models.
\subsection{Motion Vector based Alignment}
In VSR methods, alignment between neighboring frames is important for good performance. In this paper, for alignment, we warp the HR information of the past frame with MVs. Different from the interpolation filter used in H.264, the bilinear interpolation filter is applied to the pixels for efficiency if the MV is fractional. When there is an insufficient temporal connection between blocks, the video encoder utilizes intra-prediction. Since the intra-blocks mainly appear in the keyframe (the first frame of a video clip) and there are few intra-predicted blocks in most frames, for blocks with intra-prediction, we transfer the features of the same position in the adjacent frame. To a common format, we set $MV = (0,0)$ for intra-blocks. We can formulate a motion field MV with size $H\times W\times 2$ like optical flow. $H$ and $W$ are the height and width of the input LR frame, respectively. The third dimension indicates the relative position in the width and height directions. So the MV is an approximate alternative to optical flow. In this way, we bypass the complicated motion estimation. The MV-based alignment can boost the performance of existing unidirectional recurrent VSR models and even achieve comparable performance with optical flow based alignment, as demonstrated later.
\subsection{Residual Informed Sparse Processing}
As Fig. \ref{fig:MV model} shows, in the paper, we design a Residual informed sparse processing to reduce redundant computation. Residuals represent the difference between the warped frame and the current frame. The areas without Residuals indicate the current region can be directly predicted by sharing the corresponding patches from the reference frame. Therefore, Residuals can locate the areas that need to be further refined. With the guide of Residuals, we only make convolutions on the ``important" pixels. The features of the rest pixels are enhanced by aggregation with the MV-warped features from the past frame. As Fig. \ref{fig:MV model} shows, to make it robust, we adopt this sparse processing to the body (Resblocks) of the network, the head and tail Conv layers are applied on all pixels.

Benifict from motion estimation and motion compensation, we can easily predict the flat regions or regular structures like brick wall for current frame according to the contents of adjacent frames without loss (Residuals). Residuals are more likely to be introduced on complex textures. Because flat regions or regular structures take up the majority of the frame, Residuals are sparse in most scenes. Based on these characteristics, the proposed Residual informed sparse processing can significantly reduce the space-time redundancy computation while maintaining the comparable performance with baseline.
\begin{figure}
	\centering
	\includegraphics[width=0.5\linewidth]{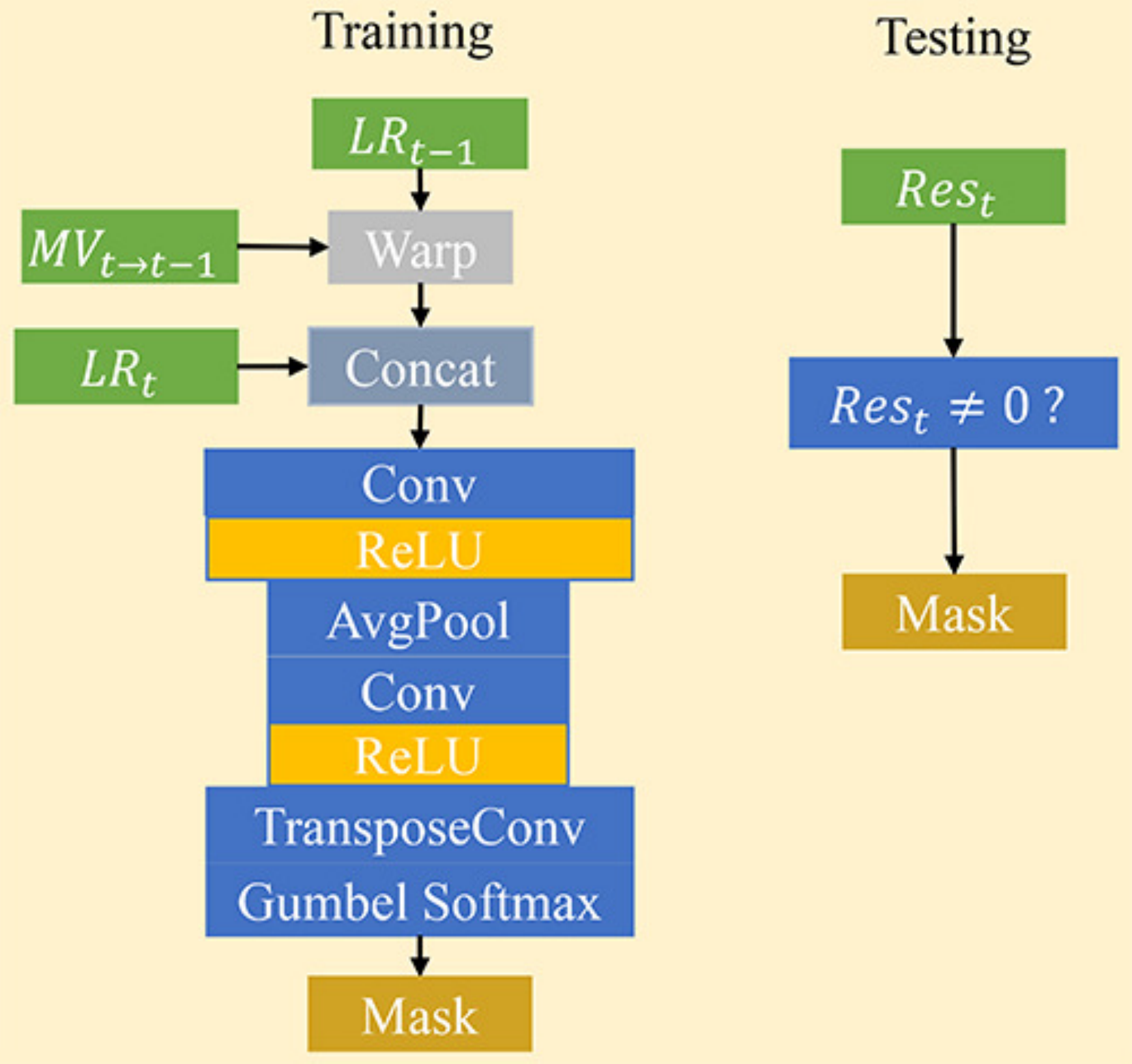}
	\caption{The sparse mask generation. $Res_t$ is the Residual extracted from compressed video. When training, we use a tiny CNN to predict a spatial mask; when testing, convolutions are only applied to pixels whose Residual is not equal to 0.}
	\label{fig:sparse model}	
\end{figure}

Because the Residuals are sparse, only a tiny part of pixels optimize the model if we directly utilize Residuals to decide where to conduct convolutions during training. In experiments, we find it hard to converge. We design a Simulated Annealing strategy to slowly reduce the number of pixels involved in training, which is a critical trick in our sparse processing. As  Fig. \ref{fig:sparse model} shows, we utilize a light CNN model to identify the changed regions according to the current frame and the MV-warped past frame. Following \cite{DBLP:journals/corr/abs-2006-09603}, Gumbel softmax trick\cite{DBLP:conf/iclr/JangGP17} is used to produce a spatial mask $M\in R^{H \times W}$ with the output features $F \in R^{2\times H \times W}$. 
\begin{equation}
	M[x,y] = \frac{exp((F[1,x,y]+G[1,x,y])/\tau)}{\sum_{i=1}^{2}exp((F[i,x,y]+G[i,x,y])/\tau)}
	\label{eq:mask train}
\end{equation}
where $x$ and $y$ are vertical and horizontal indices, $G\in R^{2\times H\times W}$ is a Gumbel noise vector with all elements following $Gumbel(0,1)$ distribution and $\tau$ is the temperature parameter. Samples from Gumbel softmax distribution become uniform if $\tau \rightarrow \infty$. When $\tau \rightarrow 0$, samples from Gumbel softmax distribution become one-hot. The predicted mask gradually becomes sparse with training. 

\noindent{\bf Training Strategy:} During training, we utilize a sparsity regularization loss to supervise the model:
\begin{equation}
	L_{reg} = \frac{1}{H\times W}\sum_{h,w}M[w,h]
	\label{eq:L mask}
\end{equation}
According the Simulated Annealing strategy, we set the weight of $L_{reg}$:
\begin{equation}
	\lambda = min(\frac{t}{T_{epoch}},1)\cdot \lambda_0
	\label{eq:L mask lamda}
\end{equation}
where $t$ is the current number of epochs, $T_{epoch}$ is empirically set to 20, and $\lambda_0$ is set to 0.004. And the temperature parameter $\tau$ in the Gumbel softmax trick is initialized as 1 and gradually decreased to 0.5:
\begin{equation}
	\tau = max(1-\frac{t}{T_{temp}},0.5)
	\label{eq:L T lamda}
\end{equation}
where $T_{temp}$ is set to 40 in this paper.

\noindent{\bf Testing:} When testing, we directly replace the mask-prediction CNN with Residuals to select the pixels to calculate. This process is formulated as:
\begin{align}
	M_{test}[x,y] = (Res[x,y]\neq 0)
	\label{eq:mask test}
\end{align}
where $Res[x,y]$ represents the Residual value at position $[x,y]$.	 When Residual is equal to 0, the pixel is skipped.
\section{Experiments}
\label{sec: experiments}
\subsection{Implementation Details}
We use dataset REDS\cite{DBLP:conf/cvpr/NahBHMSTL19} for training. REDS dataset has large motion between consecutive frames captured from a hand-held device. We evaluate the networks on the datasets REDS4\cite{DBLP:conf/cvpr/WangCYDL19} and Vid4\cite{DBLP:conf/cvpr/LiuS11}. All frames are first smoothed by a Gaussian kernel with standard deviation of 1.5 and downsampled by 4. Because our framework is designed for compressed videos, we further encode the datasets with H.264\cite{2005H}, the most common video codec, at different compression rates. The recommended CRF value in H.264 is between 18 and 28, and the default is 23. In experiments, we set CRF values to 18, 23, and 28 and use the FFmpeg codec to encode the datasets.

Our goal is to design efficient and online processing VSR systems, so we do experiments on the unidirectional recurrent VSR models. We apply our MV-based alignment to the existing models FRVSR\cite{DBLP:conf/cvpr/SajjadiVB18}, RLSP\cite{DBLP:conf/iccvw/FuoliGT19}, and RSDN\cite{DBLP:conf/eccv/IsobeJGLWT20} to verify the effect of our MV-based alignment. In the original setting, FRVSR utilizes an optical flow to align the HR output from the past frame; RLSP and RSDN do not explicitly align the information from the previous frame. For a more comprehensive comparison, we also embed a pre-trained optical flow model SpyNet\cite{DBLP:conf/cvpr/RanjanB17} into FRVSR, RLSP and RSDN to compare with our MV-based alignment. And we further fine-tune the SpyNet along with the model training. The training details follow the original works.

To evaluate the Residual informed sparse process, we first train a baseline recurrent VSR model without alignment. Then we apply MV-based alignment and Residual-based sparse processing to the baseline model to train our model. To balance model complexity and performance, the number of Resblocks for the recurrent module is set to 7. The number of feature channels is 128. We use Charbonnier loss\cite{DBLP:conf/icip/CharbonnierBAB94} as pixel-wise loss since it better
handles outliers and improves the performance over the
conventional L2-loss\cite{DBLP:conf/cvpr/LaiHA017}. The training details are provided in the supplementary material.

\subsection{Effect of MV-based Alignment}
We apply our MV-based alignment approach to the FRVSR, RLSP, and RSDN. The quantitative results are summarized in Tab. \ref{tab:MV}. XXX+Flow means that model XXX is aligned with the SpyNet. XXX+MV represents that model XXX is aligned with MVs. Original FRVSR aligns the HR estimation from the past frame by an optical flow model trained from scratch. In FRVSR+FLow, we replace the original optical flow model with pre-trained SpyNet and further refine the SpyNet when training. From the results, we can find FRVSR+Flow outperforms the original FRVSR. Probably because SpyNet estimates the optical flow more precisely than the original model. RLSP and RSDN do not explicitly align the information from the past frame. Due to the alignment, models with MV-based alignment achieve better performance than their original counterparts, even achieving comparable performance with the models with SpyNet.
And we can see that as the CRF is increased, the performance gap between optical flow-based methods and MV-based methods narrows, which makes sense since when the CRF is large, the video compression artifacts are more apparent, and the optical flow estimate mistakes are more significant. So our MV-based alignment can replace the existing optical flow estimation model in unidirectional recurrent VSR models to save computation. For RLSP and RSDN, our approach can achieve better performance with a tiny increase in runtime because of feature warping. It should be noted that our MV-based alignment does not increase the number of parameters. For FRVSR, because we remove its optical flow sub-model, our MV-based alignment can reduce the parameters and runtime but achieve superior performance over the original version. 
\renewcommand\arraystretch{1.3}
\begin{table*}[t]
	\centering
	\caption{\textbf{The quantitative comparison (PSNR/ SSIM/ LPIPS)} on REDS4\cite{DBLP:conf/cvpr/WangCYDL19}. PSNR is calculated on Y-channel; SSIM and LPIPS are calculated on RGB-channel. \textcolor{red}{Red} and \textcolor{blue}{blue} colors indicate the best and the second-best performance, respectively. 4$\times$ upsampling is performed.}
	\label{tab:MV}
	\resizebox{\textwidth}{!}{%
		\begin{tabular}{|c|ccc|c|c|}
			\hline
			\multirow{2}{*}{Model}                                                          & \multicolumn{3}{c|}{Compressed Results} & \multirow{2}{*}{Params (M)} & \multirow{2}{*}{Runtime (ms)}                                                                                            \\ \cline{2-4} 
			&  \multicolumn{1}{c|}{CRF18}                        & \multicolumn{1}{c|}{CRF23}                        & CRF28 &&                       \\ \hline
			FRVSR\cite{DBLP:conf/cvpr/SajjadiVB18}                  &  \multicolumn{1}{c|}{28.27/0.7367/0.3884}          & \multicolumn{1}{c|}{27.34/0.6965/0.4495}          & 26.11/0.6492/0.5219   &2.59&24       \\ \hline
			FRVSR+MV               &  \multicolumn{1}{c|}{\textcolor{blue}{29.01/0.7660/0.3470}} & \multicolumn{1}{c|}{\textcolor{blue}{27.77/0.7155/0.4141}} & \textcolor{blue}{26.32/0.6598/0.4969} &0.84&20 \\ \hline
			FRVSR+Flow             &  \multicolumn{1}{c|}{\textcolor{red}{29.15/0.7701/0.3393}}          & \multicolumn{1}{c|}{\textcolor{red}{27.85/0.7177/0.4076}}          & \textcolor{red}{26.32/0.6600/0.4928}   &2.28& 32      \\ \hline
			RLSP\cite{DBLP:conf/iccvw/FuoliGT19}                   &  \multicolumn{1}{c|}{28.46/0.7476/0.3614}          & \multicolumn{1}{c|}{27.47/0.7052/0.4243}          & 26.20/0.6551/0.5015   &4.37& 27      \\ \hline
			RLSP+MV                &  \multicolumn{1}{c|}{\textcolor{blue}{29.26/0.7739/0.3309}} & \multicolumn{1}{c|}{\textcolor{blue}{27.95/0.7225/0.3973}} & \textcolor{blue}{26.43/0.6646/0.4815} &4.37&28 \\ \hline
			RLSP+Flow              &  \multicolumn{1}{c|}{\textcolor{red}{29.37/0.7769/0.3249}}          & \multicolumn{1}{c|}{\textcolor{red}{28.01/0.7242/0.3947}}          & \textcolor{red}{26.44/0.6651/0.4788}   &5.81&39       \\ \hline
			RSDN\cite{DBLP:conf/eccv/IsobeJGLWT20}                   &  \multicolumn{1}{c|}{28.67/0.7575/0.3405}          & \multicolumn{1}{c|}{27.62/0.7144/0.3997}          & 26.29/0.6642/0.4731    &6.18&49      \\ \hline
			RSDN+MV                &  \multicolumn{1}{c|}{\textcolor{blue}{29.37/0.7804/0.3163}} & \multicolumn{1}{c|}{\textcolor{blue}{28.02/0.7294/0.3799}} & \textcolor{blue}{26.50/0.6724/0.4558} &6.18&51 \\ \hline
			RSDN+Flow              &  \multicolumn{1}{c|}{\textcolor{red}{29.59/0.7862/0.3094}}          & \multicolumn{1}{c|}{\textcolor{red}{28.13/0.7314/0.3770}}          & \textcolor{red}{26.51/0.6739/0.4523}   &7.62& 62      \\ \hline
		\end{tabular}%
	}
\end{table*}
\begin{figure*}[h]
	\centering
	\includegraphics[width=\linewidth]{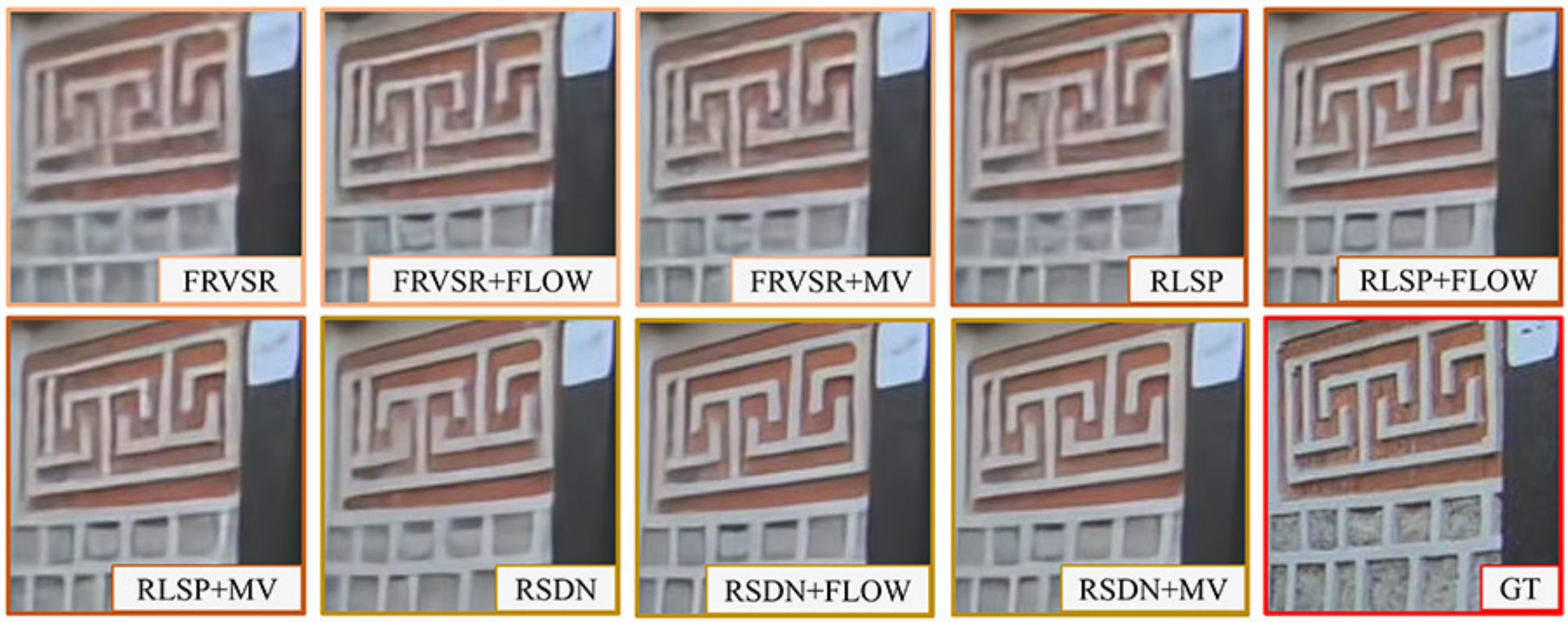}	
	\caption{Visual results on REDS4\cite{DBLP:conf/cvpr/WangCYDL19}}
	\label{com_MV}
	\centering
\end{figure*}

Fig. \ref{com_MV} shows the qualitative comparison. The models with our MV-based alignment restore finer details than the original FRVSR, RLSP, and RSDN. Compared with the models with optical flow estimation, our MV-aligned models achieve comparable visual results. More examples are provided in the Section 2.1 of supplementary material.
\begin{table*}[h]
	\centering 
	\caption{\textbf{The quantitative comparison (PSNR/ SSIM/ LPIPS)} between image alignment and feature alignment on REDS4\cite{DBLP:conf/cvpr/WangCYDL19}. PSNR is calculated on Y-channel; SSIM and LPIPS are calculated on RGB-channel. The best results are highlighted in bold.}
	\label{tab:alignment}
	\resizebox{0.8\textwidth}{!}{%
		\begin{tabular}{|c|c|c|c|}
			\hline
			Model & CRF18                        & CRF23                        & CRF28                        \\ \hline
			(a)   & 28.59/0.7546/0.3420          & 27.56/0.7122/0.3999          & 26.26/0.6622/0.4719          \\ \hline
			(b)   & \textbf{29.32/0.7783/0.3186} & \textbf{28.00/0.7273/0.3818} & \textbf{26.47/0.6706/0.4569} \\ \hline
			(c)   & 29.11/0.7675/0.3294          & 27.83/0.7172/0.3957          & 26.33/0.6604/0.4787          \\ \hline
		\end{tabular}%
	}
\end{table*}

\noindent \textbf{Image Alignment Vs Feature Alignment:}
As mentioned above, spatial alignment plays an important role in the VSR systems. The existing works with alignment can be divided into two categories: image alignment and feature alignment. We conduct experiments to analyze each of the categories and explain our design considerations about alignment. We design a recurrent baseline without alignment (Model (a)) and its MV-aligned versions. Model (b) is the MV-aligned model in feature space. And we apply MV-alignment on the HR prediction of the past frame to build a Model (c) with image alignment. The results are summarized in Tab. \ref{tab:alignment}. The models with alignment outperform the baseline model, which further demonstrates the importance of alignment. And we find Model (b) achieves better performance than Model (c), so the alignment in feature space is more effective than in pixel level. The reason is that MV is block-wise motion estimation, the warped images inevitably suffer from information distortion. But there is a certain degree of redundancy in feature space, and this phenomenon is alleviated. Besides, the features contain more high-frequency information than images.

\subsection{Effect of Residual Informed Sparse Processing}
We apply the Residual informed sparse processing to the aligned model to get a more efficient model. The quantitative results are summarized in Tab. \ref{tab:sparse}. The Baseline represents the baseline mentioned in Section 4.1; Baseline+MV means the MV-aligned model. MV+Res is the Residual-informed sparse processing. The \textbf{Sparse rate} is the ratio of pixels skipped by the network to all pixels in the image. As Tab. \ref{tab:sparse} shows, benefit from MV-based alignment, Baseline+MV achieves significant gains over the Baseline. The most gratifying result is that our sparse processing with MV-alignment and Residuals achieves a superior or comparable performance over Baseline with lots of computation saved. For the default CRF 23 in FFmpeg, our model can save about 70\% computation on REDS4 and Vid4. CRF 18 means that the encoded video is visually lossless. So it needs more Residuals to decrease the encoding error. The sparse processing can save about 50\% computation under this condition and achieve better performance than Baseline. For CRF 28, the sparse processing can save much more computation because the Residuals are sparser, and the performance is still comparable with the Baseline.
\begin{table*}[h]
	\caption{\textbf{The quantitative results (PSNR/ SSIM/ Sparse rate)} of Residual informed sparse model on REDS4\cite{DBLP:conf/cvpr/WangCYDL19} and Vid4\cite{DBLP:conf/cvpr/LiuS11}. PSNR is calculated on Y-channel; SSIM is calculated on RGB-channel. The Sparse rate is the ratio of pixels skipped by the network to all pixels in the image. \textcolor{red}{Red} and \textcolor{blue}{blue} colors indicate the best and the second-best performance, respectively. 4$\times$ upsampling is performed.}
	\label{tab:sparse}
	\resizebox{\textwidth}{!}{%
		\begin{tabular}{|c|ccc|ccc|}
			\hline
			\multirow{2}{*}{Model} & \multicolumn{3}{c|}{REDS4\cite{DBLP:conf/cvpr/WangCYDL19}}                                                                                                            & \multicolumn{3}{c|}{Vid4\cite{DBLP:conf/cvpr/LiuS11}}                                                                                                            \\ \cline{2-7} 
			& \multicolumn{1}{c|}{CRF18}                        & \multicolumn{1}{c|}{CRF23}                        & CRF28                        & \multicolumn{1}{c|}{CRF18}                        & \multicolumn{1}{c|}{CRF23}                        & CRF28                        \\ \hline
			Baseline          & \multicolumn{1}{c|}{28.59/0.7546/0.}          & \multicolumn{1}{c|}{27.56/0.7122/0.}          & \textcolor{blue}{26.26}/\textcolor{blue}{0.6622}/0.          & \multicolumn{1}{c|}{24.61/0.6668/0.}          & \multicolumn{1}{c|}{23.91/\textcolor{blue}{0.6135}/0.}          & \textcolor{blue}{22.87}/\textcolor{blue}{0.5429}/0.          \\ \hline
			Baseline+MV       & \multicolumn{1}{c|}{\textcolor{red}{29.32}/\textcolor{red}{0.7783}/0.} & \multicolumn{1}{c|}{\textcolor{red}{28.00}/\textcolor{red}{0.7273}/0.} & \textcolor{red}{26.47}/\textcolor{red}{0.6706}/0. & \multicolumn{1}{c|}{\textcolor{red}{25.13}/\textcolor{red}{0.6990}/0.} & \multicolumn{1}{c|}{\textcolor{red}{24.20}/\textcolor{red}{0.6355}/0.} & \textcolor{red}{23.01}/\textcolor{red}{0.5557}/0. \\ \hline
			
			MV+Res   & \multicolumn{1}{c|}{\textcolor{blue}{29.03}/\textcolor{blue}{0.7639}/\textcolor{red}{0.56}}          & \multicolumn{1}{c|}{\textcolor{blue}{27.72}/\textcolor{blue}{0.7131}/\textcolor{red}{0.75}}          & 26.15/0.6516/\textcolor{red}{0.89}         & \multicolumn{1}{c|}{\textcolor{blue}{25.02}/\textcolor{blue}{0.6800}/\textcolor{red}{0.49}}          & \multicolumn{1}{c|}{\textcolor{blue}{24.04}/0.6132/\textcolor{red}{0.72}}          & 22.81/0.5333/\textcolor{red}{0.90}         \\ \hline
		\end{tabular}%
	}
\end{table*}

\begin{figure*}[h]
	\centering 
	\includegraphics[width=0.9\linewidth]{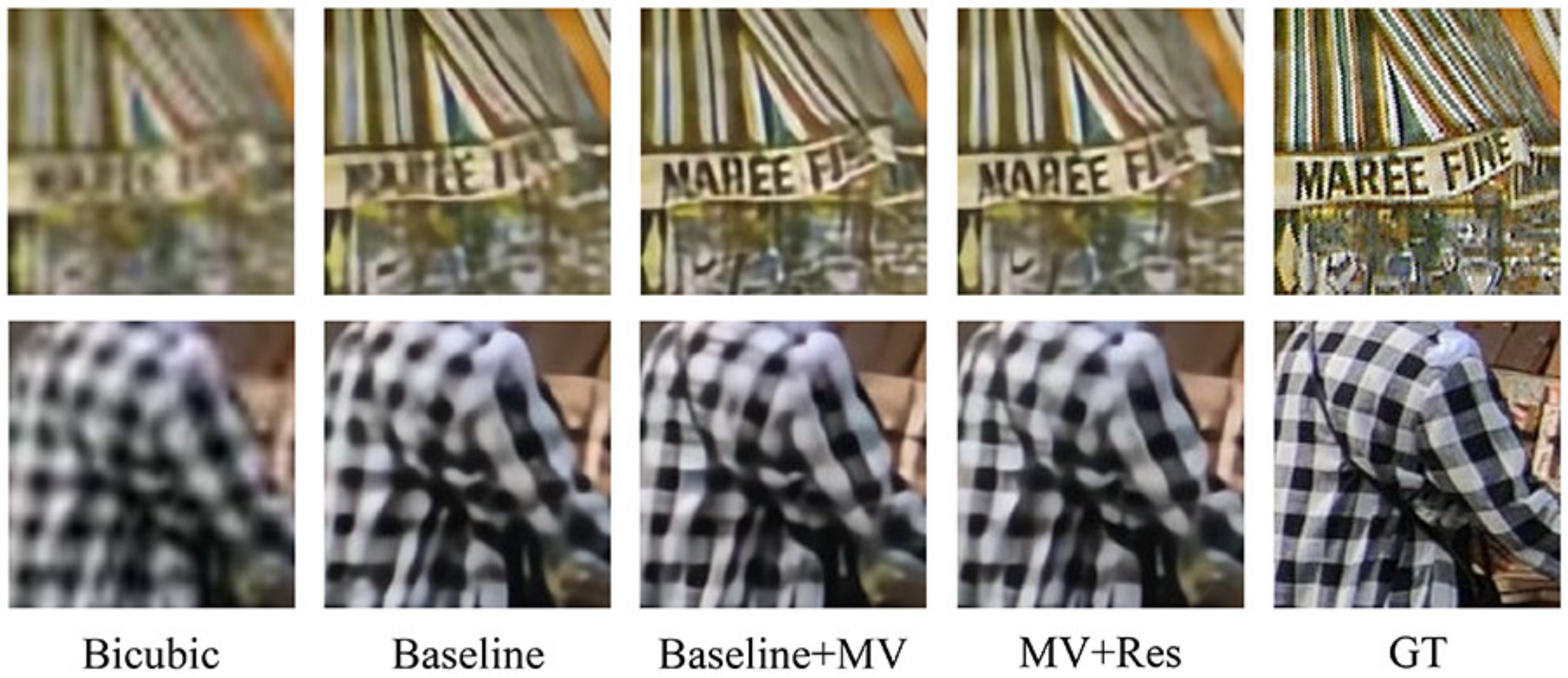}	
	\caption{Visual results of the Residual informed sparse process on Vid4\cite{DBLP:conf/cvpr/LiuS11} and REDS4\cite{DBLP:conf/cvpr/WangCYDL19}}
	\label{com_sparse}
	\centering
\end{figure*}

We conduct qualitative comparisons on datasets REDS4 and Vid4. The results are shown in Fig. \ref{com_sparse}. The Residual informed model achieves finer details than the Baseline. More examples are provided in the Section 2.2 of supplementary material.

\begin{figure*}[t]
	\centering
	\includegraphics[width=0.9\linewidth]{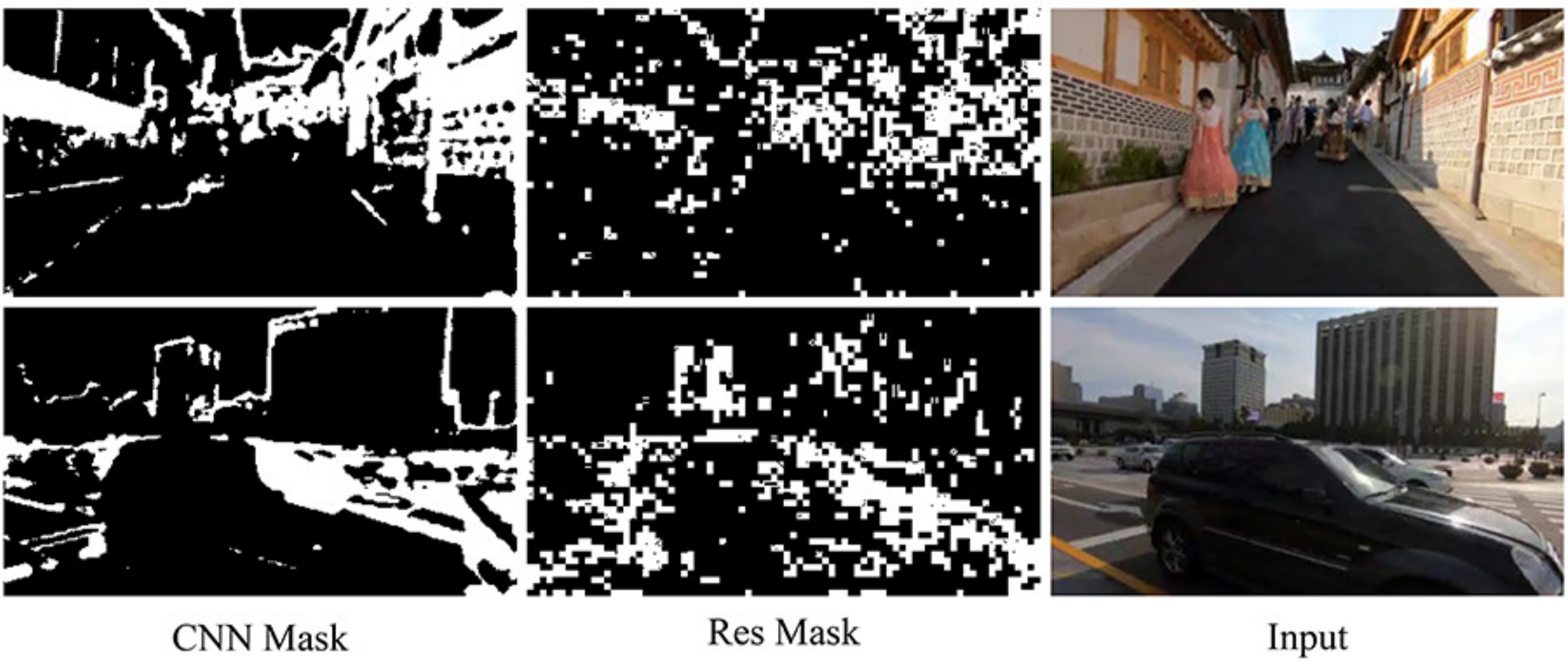}	
	\caption{Visual results of the spatial mask on REDS4\cite{DBLP:conf/cvpr/WangCYDL19}}
	\label{fig:mask}
	\centering
\end{figure*}

\noindent \textbf{CNN-based Mask Vs Residual-based Mask:}
We use a light CNN to predict the spatial mask for our Residual informed sparse processing during training. And when testing, we directly extract the Residuals from compressed videos to generate the spatial mask. In this section, we analyze the characteristics of the CNN-predicted mask and Residual-generated mask. As Fig. \ref{fig:mask} shows, we can quickly identify the contours of objects and locate the details and textures from CNN-based masks. The Residual-based masks focus on the errors between the recurrent frame and the MV-warped past frame. Because Residuals are more likely to appear in the areas with details, the highlights of Residual-based masks also follow the location of details. Besides, the CNN-based masks are more continuous than the Residual-based mask. 
We also present the performance of the models with CNN-based mask and Residual-based mask in Tab. \ref{tab:mask}. The results show that different from the Residual-based mask, the Sparse rate of CNN-base masks changes little with different CRF. So the CNN-based mask only highlights the main objects in the image. The Residual-based masks focus on the errors about MV-based alignment. For CRF 18, the information loss is slight, so the amount of Residuals is large, and the model achieves better performance than the model with the CNN-based mask. And for CRF 23 and 28, our model also outperforms the model with the CNN-based mask with a similar Sparse rate. The reason is that our Residual-based model follows the characteristics of video compression and is more suitable for models with MV-based alignment. Our Residual-based mask locates the ``important" areas that need to be refined more precisely.
\begin{table*}[h]
	\centering
	\caption{\textbf{The quantitative comparison (PSNR/ SSIM/ Sparse rate)} about spatial mask on REDS4\cite{DBLP:conf/cvpr/WangCYDL19}. PSNR is calculated on Y-channel; SSIM and LPIPS are calculated on RGB-channel. The best results are highlighted in bold.}
	\label{tab:mask}
	\resizebox{0.7\textwidth}{!}{%
		\begin{tabular}{|cc|c|c|}
			\hline
			\multicolumn{2}{|c|}{Model}                                                                                   & CNN Mask            & Res Mask            \\ \hline
			\multicolumn{1}{|c|}{\multirow{3}{*}{\begin{tabular}[c]{@{}c@{}}Compression\\  results\end{tabular}}} & CRF18 & 28.82/0.7492/\textbf{0.74} & \textbf{29.03/0.7639}/0.56 \\ \cline{2-4} 
			\multicolumn{1}{|c|}{}                                                                                & CRF23 & 27.62/0.7040/\textbf{0.76} & \textbf{27.72/0.7131}/0.75 \\ \cline{2-4} 
			\multicolumn{1}{|c|}{}                                                                                & CRF28 & 26.08/0.6456/0.79 & \textbf{26.15/0.6516/0.89} \\ \hline
		\end{tabular}%
	}
\end{table*}
\subsection{Temporal Consistency}
\begin{figure*}[h]
	\centering 
	\includegraphics[width=0.8\linewidth]{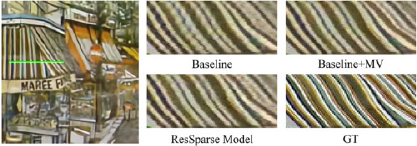}	
	\caption{Visualization of temporal profile for the green line on the calendar sequence with CRF 23.}
	\label{fig:temp}
	\centering
\end{figure*}

Fig. \ref{fig:temp} shows the temporal profile of the video super-resolution results, which is produced by extracting a horizontal row of pixels at the same position from consecutive frames and stacking them vertically. The ``ResSparse Model" is the model with our Residual informed sparse processing. The temporal profile produced by the model with our Residual informed sparse processing is temporally smoother, which means higher temporal consistency, and much sharper than the baseline model with about 70\% computation of the baseline model saved when CRF is 23.
\section{Conclusion}
\label{sec:conclusion}
This paper proposes to reuse codec information from compressed videos to assist the video super-resolution task. We employ Motion Vector to align mismatched frames in unidirectional recurrent VSR systems efficiently. Experiments have shown that Motion Vector based alignment can significantly improve performance with negligible additional computation. It even achieves comparable performance with optical flow based alignment. To further improve the efficiency of VSR models, we extract Residuals from compressed video and design Residual informed sparse processing. Combined with Motion Vector based alignment, our Residual informed processing can precisely locate the areas needed to calculate and skip the ``unimportant" regions to save computation. And the performance of our sparse model is still comparable with the baseline. Additionally, given the importance of motion information for low-level video tasks and the inherent temporal redundancy of videos, our codec information assisted framework (CIAF) has the potential to be applied to other tasks such as compressed video enhancement and denoising.

\noindent\textbf{Acknowledgement}
The authors Rong Xie and Li Song were supported by National Key R\&D Project of China under Grant 2019YFB1802701, the 111 Project (B07022 and Sheitc No.150633) and the Shanghai Key Laboratory of Digital Media Processing and Transmissions.
\clearpage
\bibliographystyle{splncs04}
%\bibliography{egbib}

\end{document}